\documentclass[11pt]{article}

\usepackage[final]{acl}

\usepackage{xcolor}
\definecolor{lightgrass}{RGB}{146,208,80}
\usepackage{algorithm}
\usepackage{algorithmic}
\usepackage{times}
\usepackage{latexsym}
\usepackage{booktabs}
\usepackage{arydshln}
\usepackage{amssymb}
\usepackage{caption}
\usepackage[most]{tcolorbox} 
\usepackage[T1]{fontenc}

\usepackage[utf8]{inputenc}

\usepackage{microtype}

\usepackage{inconsolata}

\usepackage{graphicx}

\author{Ye Mo$^{1}$\thanks{\,\, Equal contribution.}, 
Kai Ye$^{1}$\footnotemark[1],
Xianwei Mao$^{1}$\footnotemark[1], 
Zirui Shao$^{1}$,  
\textbf{Gang Huang}$^{2}$, 
\textbf{Bo Zhang}$^{3}$, \\
\textbf{Hangdi Xing}$^{1}$, 
\textbf{Kehan Chen}$^{2}$,
\textbf{Huan Zhou}$^{2}$,
\textbf{Zixu Yan}$^{2}$,
\textbf{Jiajun Bu}$^{1}$,
\textbf{Sheng Zhou}$^{1}$\thanks{\,\, Corresponding author.} \\
$^{1}$ Zhejiang Key Laboratory of Accessible Perception and Intelligent Systems, \\Zhejiang University  $^{2}$Alibaba Group, $^{3}$Shanghai AI Laboratory \\
\texttt{\{moye017, mercury0926, maoxianwei, zhousheng\_zju\}@zju.edu.cn}  }

\title{Doc-CoB: Enhancing Document Understanding with Visual Chain-of-Boxes Reasoning}

\begin{document}
\maketitle
\begin{abstract}
Document understanding aims to perform question answering and information extraction over document images, where the visual content is highly information-dense and most queries rely on only a few relevant layout regions.
However, existing methods either adopt a one-pass strategy that implicitly assumes all layouts are equally important, or focus excessively on small regions at the cost of losing critical layout information.
To address these limitations, we introduce \textbf{Doc-CoB} (Chain-of-Boxes), a simple-yet-effective framework that integrates coarse-to-fine layout-aware visual reasoning into multimodal large language models. 
Instead of directly zooming into small regions, Doc-CoB progressively focuses on query-relevant layouts while preserving global document information. 
Specifically, it first selects key layout boxes and then focuses on them for further understanding with visual prompting.
To support this paradigm, we introduce two reasoning tasks for \textit{box recognition} and \textit{box reasoning}, with an automatic pipeline that constructs \textit{249k} training samples with intermediate visual supervision.
Experiments on seven benchmarks with four popular models show that Doc-CoB significantly improves performance, demonstrating its effectiveness and wide applicability.
The code and the data are available at {https://github.com/Doc-CoB/Doc-CoB}.
\end{abstract}

\section{Introduction}
\label{sec:intro}
Document understanding~\citep{cui2021document,xu2020layoutlm} is a fundamental task that involves question answering and information extraction based on document images, such as forms, receipts, reports, and scanned pages. 
In recent years, multimodal large language models (MLLMs)~\citep{bai2025qwen3vltechnicalreport, comanici2025gemini25pushingfrontier,grattafiori2024llama3herdmodels,sensenova-si} have demonstrated promising capabilities on document understanding~\citep{hu2024mplug,ding2025survey,zhou2024doge,luo2024layoutllm,mo2025tablenarrator}, driving progress in both academic research and industrial applications.

\begin{figure*}[t]
    \centering
    \includegraphics[width=1\linewidth]{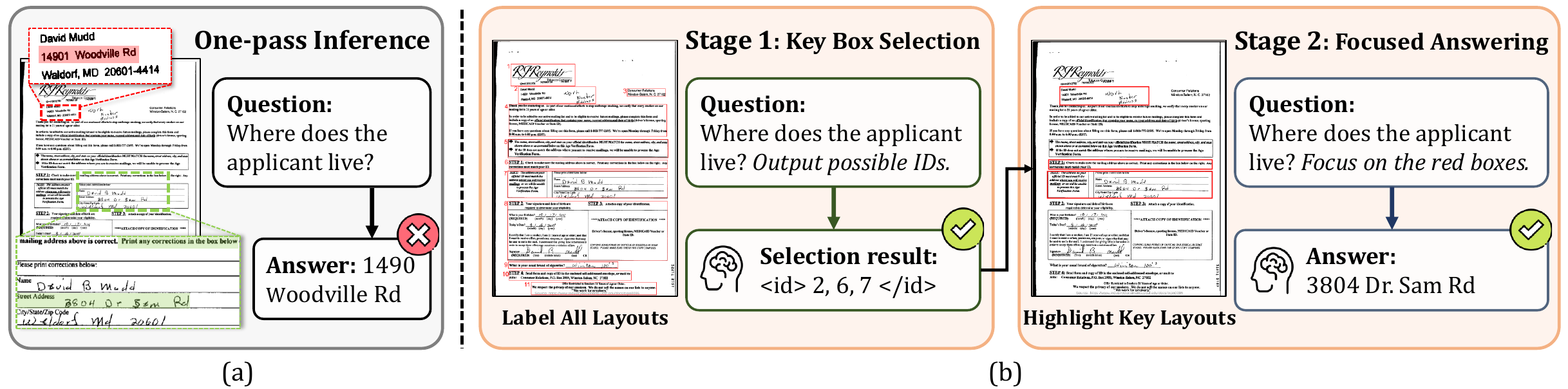}
    \caption{(a) Current MLLM attends to a wrong region and produces an incorrect answer, instead of obtaining the answer from the correct layout region below. (b) Overview of Doc-CoB. S1: Key Box Selection identifies boxes relevant to the query. S2: Focused Answering generates the final answer from the visually prompted image.}
    \label{fig:toyexample}
\end{figure*}

Most existing MLLMs generate responses to queries by processing entire images in a uniform manner \citep{chen-etal-2025-multimodal,fu2025multimodallargelanguagemodels}, implicitly assuming that all regions are equally important.
However, this overlooks the information density \citep{cui2021document} of document images, where only a few layout regions are task-relevant \citep{cao2023attention}, while the remainder is redundant.
As shown in Figure~\ref{fig:toyexample} (a), while not all regions are relevant to the query \textit{``Where does the applicant live?''}, the model is misled by excessive redundant information and infers the answer from a wrong region.

Despite the emergence of coarse-to-fine strategies in general visual QA \cite{yu2025attention}, many existing approaches have drifted toward the opposite extreme. Specifically, some works perform direct cropping or zoom-in on selected regions after a localization stage \citep{luan2024textcot, shao2024visual}. 
While such designs reduce visual redundancy, they inevitably discard layout information that is unique and critical to document images, such as spatial relationships, relative positioning, and cross-region dependencies. 
As a result, these methods may struggle with layout-sensitive queries and complex document structures.
This raises a research question: \textit{How to follow the coarse-to-fine paradigm while preserving layout information for document understanding?}

To address the above challenge, we propose \textbf{Doc-CoB} (Chain-of-Boxes), a framework that enables MLLMs to perform document understanding through coarse-to-fine, layout-aware visual reasoning without modifying their architecture. 
We decompose the process into two stages: \textit{(S1) Key Box Selection} and \textit{(S2) Focused Answering}, as illustrated in Figure~\ref{fig:toyexample} (b). 
Both stages employ visual prompting \citep{yang2024fine, shtedritski2023does} to explicitly guide the model’s attention to key layout (box) while suppressing visual redundancy.
We then introduce two reasoning tasks: \textit{(T1) Box Recognition} and \textit{(T2) Box Reasoning}, which aim to improve the model’s understanding of document layout and its associated semantic information.
Finally, to address the challenge of limited training data, we construct a large-scale dataset with intermediate visual supervision tailored to Doc-CoB. Built upon an automatic data generation pipeline, the dataset has 249K training samples collected from nine document datasets, where layout structure and semantic roles are jointly encoded.

Extensive experiments and ablation studies examine Doc-CoB from multiple complementary perspectives.
All MLLMs achieve improved performance when integrated with Doc-CoB, with an 8B model even surpassing GPT-4o. 
Further in-depth analysis reveals that the explicit Doc-CoB reasoning paradigm and the proposed reasoning tasks are both critical.
Beyond in-domain evaluation, Doc-CoB exhibits strong generalization, remaining effective under out-of-domain settings.
Our contributions are as follows:
\begin{itemize}
    \item We introduce Doc-CoB, a framework that enables MLLMs to perform document understanding through coarse-to-fine, layout-aware visual reasoning. 
    \item We propose two reasoning tasks to further enhance the reasoning capabilities of MLLMs. In addition, we construct a dataset of 249k samples across nine document datasets, each annotated with intermediate visual supervision tailored for Doc-CoB. 
    \item Doc-CoB improves performance on seven benchmarks, surpasses strong baselines, and remains compatible with various MLLM architectures and scales. 

\end{itemize}

\section{Related Work}
\label{sec:related}
\textbf{Multimodal Reasoning.} A critical step in advancing AI toward human-level intelligence is enabling the transition from basic perception to complex reasoning~\cite{wang2025multimodal}.
Recent works~\cite{jaech2024openai,guo2025deepseek} have shown that incorporating reasoning can significantly enhance model performance.
One widely used class of methods is Chain-of-thought (CoT) \citep{wei2022chain}, which encourages large language models (LLMs) to clarify reasoning, specifically by adding thinking processes \citep{kang2025c3ot,li2025structured,miao2024chain,mo2026open}. 

Meanwhile, progress has been made in reasoning for MLLMs. 
MM-CoT \citep{zhang2023multimodal} proposed a two-stage reasoning framework by first generating rationales and then generating answers.
Subsequent works \citep{he2024multi,wang2024t} further improve the design on vision-language fusion mechanisms to automate the construction of CoT. 
In recent years, visual prompts in MLLMs have attracted significant attention \citep{xu2025progressive,parmar2025object}. 
Researchers \citep{luan2024textcot} propose pioneering work to implement visual prompt for visual question answer task. However, the text form output of MLLMs significantly constrains accuracy of the prompt.

\textbf{MLLMs for Document Understanding.}
Document understanding \citep{zhang2024token,luo2025bi} is an increasingly prominent research area driven by the growing industrial demand for efficient information processing. 
Recent advancements \citep{chen2024internvl,qwen2.5} in general MLLMs have enhanced the encoding resolution of document images, markedly improving the efficacy of document understanding tasks. A number of MLLMs have been specifically developed to tackle challenges in document understanding. For instance, mPLUG-DocOwl series \citep{ye2023mplug,hu2024mplug} unify task processing across five types of document images, achieving notable performance in document comprehension. 
However, despite the significant progress made by existing MLLMs, some findings indicate that these models do not focus on relevant areas when responding to queries \citep{yu2025attention}, 
especially problematic for complex documents.

\textbf{Layout-aware Document Understanding.}
Recent studies incorporate document layout information from several complementary perspectives.
GeoLayoutLM~\citep{luo2023geolayoutlm} models geometric relationships among layout entities.
Fox~\citep{liu2024focus} improves fine-grained regional perception by enabling MLLMs to operate on user-specified document regions through positional visual prompts.
LayoutLLM~\citep{luo2024layoutllm} moves further toward explicit document reasoning by decomposing document understanding into multiple reasoning steps and representing the reasoning process with textual CoT.
Unlike prior methods that focus on layout representation, regional perception, or textual reasoning, Doc-CoB explicitly models query-conditioned layout relevance, and reasons over the layouts that contribute to the downstream task.

\section{Methodology}

The Doc-CoB framework (Algorithm~\ref{alg:doc-cob}) is designed to address the challenge of balancing key layout regions and global contextual information in document understanding.
Similar to human document reading behavior, document processing usually starts with segmenting the layout into ``boxes'', and performing knowledge-guided saccadic scanning to select relevant boxes for focused understanding.
Inspired by this reading behavior, we define Doc-CoB as a two-stage paradigm: \textit{S1: Key Box Selection} and \textit{S2: Focused Answering}. 

\subsection{S1: Key Box Selection}
\label{sec:reasoning_definition}
Given a document image $x_I$ and a query $Q$, Doc-CoB uses a layout analyzer $\Lambda(\cdot)$ to generate a set of candidate layout boxes \(B = \Lambda(x_I)\),
where each box $b \in B$ is assigned a unique index $id$.
Then we construct a visually prompted image $X_{I}^{\mathrm{S1}}$ by overlaying all candidate boxes $B$ on $x_I$, and labeling each box with its index $id$. Then the MLLM $f(\cdot)$ selects the subset of boxes $B^{\mathrm{key}}$that are most relevant to $Q$:
\begin{equation}
\mathcal{I}_{\text{key}}
= f\!\bigl(\,[X_{I}^{\mathrm{S1}}; B],\, Q,\, P_{\mathrm{S1}} \bigr),
\end{equation}
\begin{equation}
B^{\mathrm{key}} = \{\, b \in B \mid id(b) \in \mathcal{I}_{\text{key}} \,\}.
\end{equation}
where $\mathcal{I}_{\text{key}} \subseteq \{1,\dots,|B|\}$, and $P_{\mathrm{S1}}$ is the prompt used in the key box selection stage (see Appendix~\ref{ap2}).
In practice, we use a layout analyzer and formulate this stage as a multiple-choice prediction problem over the candidate boxes.

\subsection{S2: Focused Answering.}
After obtaining $B^{\mathrm{key}}$, we perform inference by directing the MLLM’s attention to the selected boxes. 
Specifically, we generate another visually prompted image $X_{I}^{\mathrm{S2}}$ by marking the selected boxes with red borders, without altering the remaining visual content. 
This design preserves the global document context while emphasizing relevant layout regions,
which is crucial for layout-dependent queries (e.g., \textit{``What is the first bullet on the right side?''}). 
The final answer $A$ is produced by
\begin{equation}
A
= f\!\bigl(\,[X_{I}^{\mathrm{S2}}; B^{\mathrm{key}}],\, Q,\, P_{\mathrm{S2}} \bigr),
\end{equation}
where $P_{\mathrm{S2}}$ is a short instruction \textit{``Please pay more attention to the red boxes''}.

\begin{figure*}[h]
    \centering
    \includegraphics[width=1\linewidth]{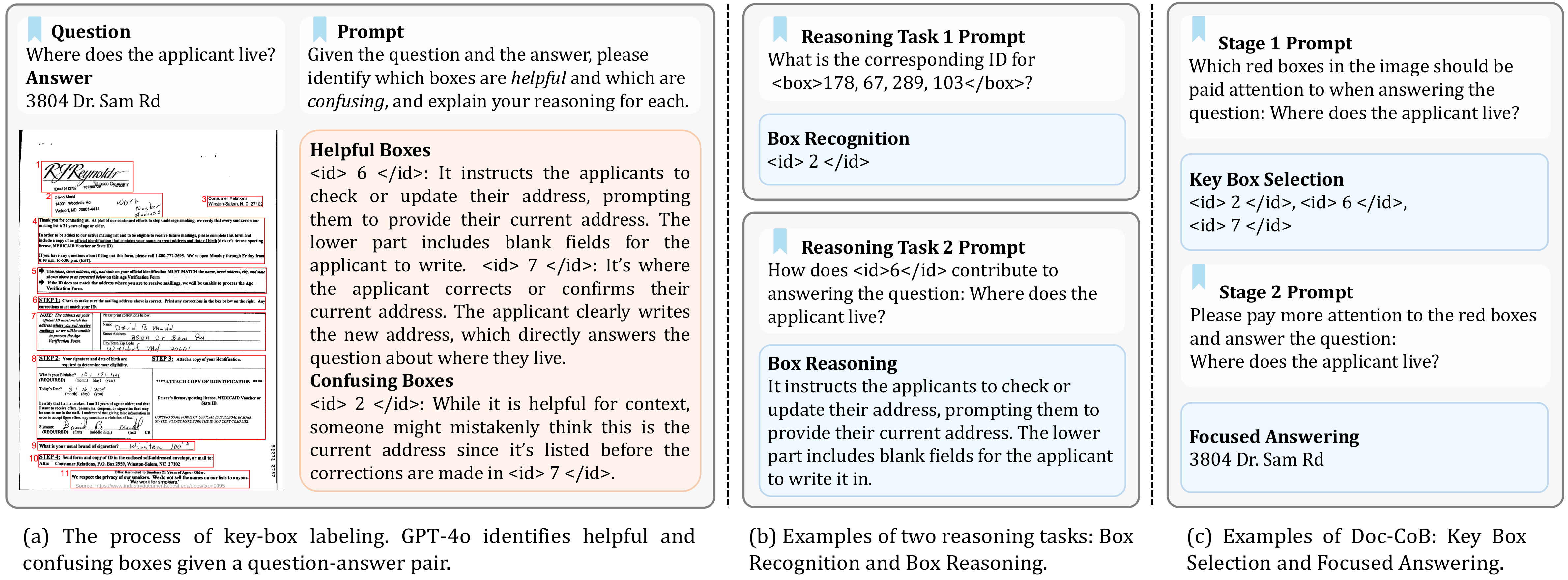}
    \caption{
Training data generation pipeline and examples of two reasoning tasks and Doc-CoB training data.}
    \label{fig:mainfigure}
\end{figure*}

\begin{algorithm}[tb]
\small
\caption{Doc-CoB Algorithm}
\label{alg:doc-cob}
\textbf{Input}: Document image $x_I$ and question $Q$ \\
\textbf{Output}: Answer $A$
\begin{algorithmic}[1]

\STATE $B \leftarrow \Lambda(x_I)$
\hfill $\triangleright$ \textit{Layout analyzer extracts candidate boxes}

\STATE $X_I^{S1} \leftarrow \textsc{RenderID}(x_I, B)$
\hfill $\triangleright$ \textit{Overlay each box with a unique ID}

\STATE $\mathcal{I}_{\text{key}} \leftarrow \textsc{SelectIDs}\big(
f([X_I^{S1}; B], Q, P_{S1})\big)$
\hfill $\triangleright$ \textit{Stage 1: select key box IDs}

\STATE $B_{\text{key}} \leftarrow \{ b_i \mid i \in \mathcal{I}_{\text{key}} \}$
\hfill $\triangleright$ \textit{Retrieve key boxes by IDs}

\STATE $X_I^{S2} \leftarrow \textsc{DrawBox}(x_I, B_{\text{key}})$
\hfill $\triangleright$ \textit{Highlight selected boxes}

\STATE $A \leftarrow f([X_I^{S2}; B_{\text{key}}], Q, P_{S2})$
\hfill $\triangleright$ \textit{Stage 2: generate answer conditioned on key boxes}

\STATE \textbf{return} $A$

\end{algorithmic}
\end{algorithm}

\subsection{Reasoning Tasks}
\label{sec:enabling_tasks}

Beyond proposing a coarse-to-fine framework tailored for document, we also need to enhance the model’s visual and semantic understanding of layouts.
However, we observe that the intrinsic capabilities of existing MLLMs are still insufficient to fully support the Doc-CoB framework. 
First, the model may fail to align a box $b$ with its $id$ in the visually prompted image. 
Second, even when the box–id mapping is correct, the model frequently struggles to identify the semantic role of each layout box when answering the query.

To address these limitations, we introduce two reasoning tasks: \textit{Box Recognition} and \textit{Box Reasoning}. 
In the Box Recognition task, the MLLM is given $X^{\mathrm{S1}}$ and a box $b_i$ and is prompted, \textit{``What is the index of ${b_i}$?''}. It should return the corresponding $id_i$. 
In Box Reasoning task, the model is given $X^{\mathrm{S1}}$, a question $Q$, and an index $id_i$, and is prompted, \textit{``What role does ${id_i}$ play in answering ${Q}$?''}. It should describe the role of $b_i$ with respect to $Q$.
These two reasoning tasks inject layout awareness and query-conditioned reasoning into the model, thereby supporting Doc-CoB reasoning. 
We construct the training samples for these two reasoning tasks based on nine datasets (Sec.~\ref{sec:data_generation}).

\subsection{Training Dataset Generation}
\label{sec:data_generation}

Existing document understanding corpora lack the annotations required by Doc-CoB. To address this, we construct an automatic data generation pipeline (Figure \ref{fig:mainfigure}) that uses GPT-4o \citep{gpt4o} as the linguistic annotator and MinerU \cite{wang2024mineruopensourcesolutionprecise}, a layout analysis tool, as the visual annotator. 
Starting from widely used document datasets, we retain their images $X_I$, question $Q$ and answer $A$.
For each $X_I$, we apply MinerU to obtain a set of layout boxes $B=\{b_i\}_{i=1}^{|B|}$. These boxes serve as the visual primitives for subsequent reasoning.

The key annotation step is to identify the subset of key boxes $B^{\text{key}} \subseteq B$. 
Given $(X_I^{\text{S1}}, B, Q, A)$, GPT-4o returns two disjoint subsets: the \textit{helpful box} $B^H$ and the \textit{confusing box} $B^C$, such that $B^{\text{key}} = B^H \cup B^C$. It also generates a natural language description $D(b)$ for each $b \in B^{\text{key}}$, explaining its function in answering $Q$. Boxes in $B^H$ contain the useful evidence for $A$, while boxes in $B^C$ resemble the query in wording, format, or position but do not contain the answer.

We design a prompt that specifies the task requirements, defines helpful and confusing boxes, and outlines the output format to ensure consistent annotations. Data quality is maintained through automatic sanity checks, including verification that $A$ is entailed in $B_H$, as well as targeted manual review. Details of the prompt and quality assurance process are provided in the Appendix \ref{ap1}.

Specifically, we select nine widely used document datasets: DocVQA \citep{mathew2021docvqa}, DUDE \citep{van2023document}, DeepForm \citep{borchmann2021due}, FUNSD \citep{jaume2019funsd}, SROIE \citep{huang2019icdar2019}, VRDU-Ad-Buy \citep{Wang_2023}, VRDU-Registration-Form \citep{Wang_2023}, FeTaQA \citep{nan2022fetaqa}, and PubLayNet \citep{zhong2019publaynet}. We directly adopt the question–answer pairs provided by DocVQA and DUDE. We convert the annotations of FUNSD, SROIE, FeTaQA, and PubLayNet into question–answer pairs following \citet{luo2024layoutllm}, and those of DeepForm, VRDU-Ad, and VRDU-RF following \citet{hu2024mplug}.
As a result, we construct the training datasets for Doc-CoB (249,601 samples), Box Recognition (126,970 samples), and Box Reasoning (269,108 samples). Details are in the Appendix \ref{ap3}.

\section{Experiments}
\label{sec:experiments}

\begin{table*}[t]
\begin{center}
\resizebox{\textwidth}{!}{
\begin{tabular}{@{}lccccccc@{}}
\toprule
 & DeepForm & SROIE & FUNSD & DUDE & VRDU-Ad & VRDU-RF & DocVQA \\ 
\midrule
Qwen2.5-VL-3B* & 38.20 & 94.99 & 75.94 & 63.88 & 68.30 & 82.43 & 90.63 \\
Qwen2.5-VL-3B-R & 42.11 & 95.52 & 77.50 & 62.73 & 69.99 & 82.79 & 90.79 \\
\textbf{Qwen2.5-VL-3B-CoB} & \textbf{\textbf{79.20}} & \textbf{96.07} & \textbf{80.55} & \textbf{64.36} & \textbf{88.66} & \textbf{90.83} & \textbf{91.11} \\
\addlinespace[2pt]
\hdashline
\addlinespace[2pt]
DocOwl 1.5* & 74.98 & 95.67 & 73.33 & 57.21 & 88.02 & 85.61 & 81.53 \\
DocOwl 1.5-R & 77.86 & 96.35 & 76.00 & 57.52 & 89.71 & 86.54 & 82.02 \\
\textbf{DocOwl 1.5-CoB} & \textbf{79.33} &
\textbf{96.52} &
\textbf{81.77} &
\textbf{57.87} &
\textbf{93.35} &
\textbf{90.92} &
\textbf{82.88} \\
\addlinespace[2pt]
\hdashline
\addlinespace[2pt]
InternVL2-2B* & 34.15 & 87.11 & 73.79 & 58.84 & 54.94 & 63.09 & 85.04 \\
InternVL2-2B-R & 48.49 & 88.02 & 72.33 & 58.31 & 70.94 & 73.94 & 85.27 \\
\textbf{InternVL2-2B-CoB} & \textbf{78.57} & \textbf{95.94} & \textbf{81.87} & \textbf{59.92} & \textbf{92.63} & \textbf{90.70} & \textbf{85.61} \\
\addlinespace[2pt]
\hdashline
\addlinespace[2pt]
InternVL2-8B* & 36.33 & 90.39 & 75.17 & 64.12 & 58.66 & 67.58 & 90.27 \\
InternVL2-8B-R & 48.49 & 94.31 & 73.91 & 63.42 & 77.77 & 80.99 & 91.00 \\
\textbf{InternVL2-8B-CoB} & \textbf{\underline{80.06}} &
\textbf{\underline{97.18}} &
\textbf{\underline{82.95}} &
\textbf{\underline{65.93}} &
\textbf{\underline{93.76}} &
\textbf{\underline{92.64}} &
\textbf{91.16} \\

\midrule
\addlinespace[0.5pt]
\multicolumn{8}{c}{\small\textit{\textbf{Much Larger MLLMs}}}\\[-1pt]
InternVL2-40B & 43.71 & 92.03 & 75.87 & 65.80 & 64.81 & 69.17 & \underline{93.86} \\
GPT-4o  & 44.75 & 91.87 & 80.21 & 65.57 & 78.90 & 72.96 & 91.05 \\

\midrule
\addlinespace[0.5pt]
\multicolumn{8}{c}{\small\textit{\textbf{Visual Prompting Methods}}}\\[-1pt]
TextCoT  & 42.45 & 81.62 & 58.60 & 43.68 & 54.58 & 59.10 & 68.29 \\
Visual CoT   & 44.76 & 88.78 & 74.90 & 53.41 & 69.14 & 77.98 & 81.16 \\

\bottomrule
\end{tabular}
}
\caption{Model performance (\%).
\textbf{Bold} numbers denote the best within each group. 
\underline{Underline} numbers indicate the best per column.
* indicates SFT baseline; -R adds reasoning tasks, and -CoB denotes models with Doc-CoB. 
}
\label{tab:mainresults}
\end{center}
\end{table*}

\begin{table*}[h]
\centering
\setlength\tabcolsep{3pt}
\resizebox{\textwidth}{!}{
\begin{tabular}{cccccccccccc}
\toprule
\# & R1 & R2 & CoB SFT & CoB Paradigm & DeepForm & SROIE & FUNSD & DUDE & VRDU-Ad & VRDU-RF & DocVQA \\ 
\midrule
1 &  &  &  &  & 36.33 & 90.39 & 75.17 & 64.12 & 58.66 & 67.58 & 90.27 \\
2 & \checkmark & \checkmark & & & 48.49 & 94.31 & 73.91 & 63.42 & 77.77 & 80.99 & 91.00 \\
3 &  &  & \checkmark & \checkmark & 64.17 & 96.37 & 79.21 & 64.32 & 83.39 & 84.44 & 91.09 \\
4 &  & \checkmark & \checkmark & \checkmark & 48.55 & 93.65 & 71.82 & 63.78 & 74.66 & 80.91 & 89.16 \\
5 & \checkmark &  & \checkmark & \checkmark & 45.82 & 91.21 & 72.51 & 63.34 & 72.97 & 77.29 & 90.79 \\ 
6 & \checkmark & \checkmark & \checkmark & & 72.17 & 96.50 & 80.76 & 64.88 & 87.29 & 85.26 & 91.12 \\
\midrule
7 & \checkmark & \checkmark & \checkmark & \checkmark & 80.06 & 97.18 & 82.95 & 65.93 & 93.76 & 92.64 & 91.16 \\
\bottomrule
\end{tabular}}
\caption{Ablation study based on InternVL2-8B. R1, R2 and CoB SFT mean the models are supervised fine-tuned on these training datasets. CoB Paradigm means the models adopt the two-stage CoB paradigm during inference.}
\label{tab:ablation_new}
\end{table*}

In this section, we investigate four research questions to evaluate the effectiveness and scalability of Doc-CoB through experimentation:

\noindent\textbf{Model performance.} Can Doc-CoB improve document understanding performance? (Sec. \ref{sec:mainres})

\noindent\textbf{Ablation analysis.} Verify the contribution of each component. (Sec. \ref{aba_experi}, Sec. \ref{keybox})

\noindent\textbf{Generalization.} Can Doc-CoB generalize to unseen domains and settings? (Sec. \ref{zero}, Sec. \ref{sec:layout_analyzers})

\noindent\textbf{Efficiency and practicality.} What is the accuracy–efficiency trade-off when integrating Doc-CoB into existing MLLMs? (Sec. \ref{tradeoff})

\subsection{Experimental Setup}
\subsubsection{Datasets and Metrics}
We evaluate Doc-CoB on seven widely used datasets, categorized into \textbf{Document QA} (DocVQA \citep{mathew2021docvqa}, DUDE \citep{van2023document}) and \textbf{Document IE} (DeepForm \citep{borchmann2021due}, FUNSD \citep{jaume2019funsd}, SROIE \citep{huang2019icdar2019}, VRDU-Ad-Buy, and VRDU-Registration-Form \citep{Wang_2023}). Detailed statistics of the test data are provided in the Appendix \ref{ap4}.  
For evaluation metrics, we use Average Normalized Levenshtein Similarity (ANLS) \citep{biten2019icdar} to assess text similarity on DocVQA, DUDE, FUNSD, and SROIE, following \citet{luo2024layoutllm}. For DeepForm, we report F1 scores following \citet{hu2024mplug}. For VRDU, we adopt micro F1 with type-aware fuzzy matching, as proposed by \citet{Wang_2023}, to accommodate minor variations in answer formats.

\subsubsection{Implementation Details.}
\label{sec:Implementation}
We select four MLLMs with varying sizes and architectures: 
InternVL2-2B \citep{chen2024far}, InternVL2-8B, Qwen2.5-VL-3B \citep{qwen2.5}, and DocOwl1.5-8B \citep{hu2024mplug}.
The InternVL2 series and Qwen2.5-VL-3B are general-purpose models, and DocOwl1.5-8B is an MLLM focusing on document understanding.

During training, we adopt a two-stage training strategy: first, we train the models on two reasoning tasks (Sec.~\ref{sec:enabling_tasks}, 396k samples), followed by training on Doc-CoB dataset (Sec.~\ref{sec:data_generation}, 249k samples). Throughout training, the visual encoder of each model remains frozen, and only the language model is updated. InternVL2 series, Qwen2.5-VL, and DocOwl1.5 are trained with learning rates of $1\mathrm{e}{-6}$, $1\mathrm{e}{-5}$, and $2\mathrm{e}{-5}$, respectively, for one epoch on 8 NVIDIA A100 GPUs. 
All randomness-inducing hyperparameters are disabled during inference to ensure consistent results. For a fair comparison, we fine-tune each model on the training set of the datasets, and report these results as SFT baselines.

\begin{table*}
\setlength\tabcolsep{4pt}
\centering
\resizebox{\textwidth}{!}{
\begin{tabular}{lcccccccc}
\toprule
& DeepForm & SROIE & FUNSD & DUDE & VRDU-Ad & VRDU-RF & DocVQA & \vline\: Avg. \\
\midrule
Qwen2.5-VL-CoB & 98.13 & 99.06 & 96.18 & 91.25 & 97.43 & 96.88 & 92.71 & \:\,\vline\: 95.95 \\
DocOwl1.5-CoB & 98.28 & 99.11 & 98.43 & 91.36 & 97.74 & 96.67 & 93.13 & \:\,\vline\: 96.39 \\
InternVL2-2B-CoB & 98.43 & 98.86 & 98.13 & 90.87 & 96.30 & 97.15 & 92.66 & \:\,\vline\: 96.06 \\
InternVL2-8B-CoB & 99.01 & 99.57 & 99.23 & 93.42 & 98.08 & 97.79 & 95.82 & \:\,\vline\: 97.56 \\ 
\bottomrule
\end{tabular}}
\caption{Performance on S1: Key Box Selection. }
\label{tab:cob_s1_f1}
\end{table*}

\subsection{Main Results}
\label{sec:mainres}
Table~\ref{tab:mainresults} presents the evaluation results.
The results show that Doc-CoB consistently improves performance across all seven datasets and all four MLLMs.
This demonstrates that Doc-CoB is a general-purpose solution applicable to a wide range of model sizes and architectures.
Notable gains appear on DeepForm, SROIE, FUNSD, and VRDU.
For instance, InternVL2-8B-CoB achieves 80.06\% on DeepForm, exceeding the 36.33\% SFT baseline by more than 43 percentage points.
We also present results for variants trained only on the reasoning tasks and evaluated with one-pass inference (-R). These settings still yield clear gains, indicating that the two tasks inject layout awareness and semantic reasoning into the MLLM and thus contribute independently to document-understanding performance.

\textbf{Comparison with Much Larger MLLMs.} We further compare with larger MLLMs, namely the leading open-source model InternVL2-40B and the leading closed-source model GPT-4o (prompts are provided in the Appendix \ref{ap6}).
With the support of Doc-CoB, the smaller InternVL2-8B-CoB achieves competitive or superior performance: it surpasses InternVL2-40B on six datasets and outperforms GPT-4o on all seven benchmarks, underscoring the effectiveness of our method.

\textbf{Comparison with Visual Prompt Methods.} Several studies investigate visual prompting and demonstrate its effectiveness on natural scene vision tasks, such as Visual CoT \citep{shao2024visual} and TextCoT \citep{luan2024textcot}. 
Although effective in natural scenes, these methods have not been explored for document understanding. 
To address this gap, we implement Visual CoT and TextCoT using InternVL2-8B.
Training and inference procedures strictly follow Sec.~\ref{sec:Implementation}.
As shown in Table~\ref{tab:mainresults}, these methods underperform on all benchmarks. We attribute this performance gap to inaccurate ROI grounding, which fails due to neglecting the critical role of document layout.

\subsection{Ablation Study}
\label{aba_experi}

\begin{table*}[h]
\begin{center}
\resizebox{\textwidth}{!}{
\begin{tabular}{lccccccc}
\toprule
& DeepForm & SROIE & FUNSD & DUDE & VRDU-Ad & VRDU-RF & DocVQA \\
\midrule
\addlinespace[0.5pt]
\multicolumn{8}{c}{\small\textit{\textbf{Model Performance $\uparrow$}}}\\[-1pt]
Cropped Image & 66.06 & 91.76 & 70.29 & 57.89 & 80.26 & 77.38 & 82.44 \\
Whole Image & 80.06 & 97.18 & 82.95 & 65.93 & 93.76 & 92.64 & 91.16 \\
\midrule
\addlinespace[0.5pt]
\multicolumn{8}{c}{\small\textit{\textbf{Visual Token Usage $\downarrow$}}}\\[-1pt]
Cropped Image & 1483.15 & 1292.76 & 1535.67  & 918.93  & 1391.12  & 1592.89 & 3448.78  \\
 & (60.09\%) &(86.88\%)  &(70.96\%) &(70.40\%) &(55.73\%) & (73.90\%) & (75.95\%) \\
Whole Image & 2468.80 & 1487.00 & 2163.34 & 1304.20 & 2496.32 & 2154.14 & 4540.47 \\
\bottomrule
\end{tabular}}
\caption{Comparison of model performance and visual token usage for full-image and cropped inputs in Stage 2.}
\label{tab:s2_input_ablation}
\vspace{-7pt}
\end{center}
\end{table*}

\textbf{Doc-CoB Design.}
Table~\ref{tab:ablation_new} reports the ablation experiment results of the InternVL2-8B.
We evaluate a one-pass variant of InternVL2-8B-CoB (\#6), which is trained with the CoB dataset but answers directly without the two-stage paradigm. While (\#6) surpasses all other baselines (\#1\textasciitilde\#5) across all benchmarks, it consistently underperforms the complete CoB (\#7),
indicating that the performance gain primarily stems from the change in reasoning paradigm rather than the training data alone.

Although the results in Table~\ref{tab:mainresults} demonstrate the benefits of adding two reasoning tasks, we further investigate a setting where the reasoning tasks are removed. The performance gap between (\#3) and the full CoB (\#7) suggests that the intrinsic reasoning capabilities of current MLLMs are still insufficient to fully exploit the potential of Doc-CoB.
The results show that Doc-CoB’s full effectiveness arises from the combination of explicit reasoning paradigm and targeted supervision.
 
\textbf{Impact of Global Context.} 
To demonstrate the necessity of preserving global context in Stage 2,
we further evaluate a cropping variant that replaces the whole-page input with the cropped area. 
As shown in Table~\ref{tab:s2_input_ablation}, cropping consistently degrades InternVL2-8B-CoB across all seven datasets: the drop ranges from 5.42 to 15.26 points, and exceeds 8 points on six datasets. 

This suggests that the selected area is insufficient when treated as an isolated snippet. 
Many document visual question answer tasks rely on global spatial context, such as references to ``the middle table'', ``the last image'', or ``the third paragraph in the right column''. 
Cropping removes these layout cues and disrupts spatial relationships across regions. 
Table~\ref{tab:s2_input_ablation} also shows that cropping reduces visual tokens by 13.06\%--44.27\%, but still retains 55.73\%--86.88\% of the original tokens. Therefore, cropping provides only moderate computational savings relative to its substantial accuracy loss, making the visually prompted whole-image input a better trade-off for Stage 2.

\subsection{Key Box Selection Performance}
\label{keybox}
Given that Doc-CoB adopts a two-stage approach, we isolate Stage 1, Key Box Selection, to assess its impact on the overall performance.  
As defined in Sec.~\ref{sec:reasoning_definition}, a ``helpful box'' contains evidence useful for answering a question.  
Therefore, we report the model's F1 score for selecting helpful boxes, denoted as $F1_H$.
As shown in Table~\ref{tab:cob_s1_f1}, on datasets where Doc-CoB achieves substantial performance improvements (such as DeepForm, SROIE and VRDU), the average $F1_H$ consistently exceeds 97\%.

In contrast, on datasets where Doc-CoB has a less pronounced impact, such as DUDE, the $F1_H$ is relatively lower.
This suggests that high-quality key-box selection tends to correlate with improved overall performance, although Table~\ref{tab:mainresults} also shows that Doc-CoB can still offer benefits even when $F1_H$ is relatively modest.
This suggests that high-quality intermediate visual reasoning effectively supports document understanding.

\subsection{Zero-shot Performance}
\label{zero}
To further validate the cross-domain scalability of our method, we conduct zero-shot evaluations of InternVL2-8B-CoB on InfographicVQA \cite{mathew2022infographicvqa} and MP-DocVQA \cite{tito2023hierarchicalmultimodaltransformersmultipage}.
For MP-DocVQA, we numbered the boxes sequentially and
fed all pages into the model for box selection. Then, we removed unselected pages, generated visual prompts based on the selected boxes, and asked the model to answer the question.
Since multiple images were input simultaneously, we had to reduce the resolution hyperparameter to avoid OOM errors. In this setting, the advantage of CoB is that by emphasizing key areas, it can offset the model’s need for high effective resolution, thereby improving performance on multi-page data.
The results demonstrate that our model can also be extended to infographics and multi-page documents.

\begin{table}
\begin{center}
\setlength\tabcolsep{4pt}
\resizebox{\columnwidth}{!}{
\begin{tabular}{lcc}
\toprule
& \multicolumn{1}{l}{InfographicVQA} & \multicolumn{1}{l}{MP-DocVQA} \\ 
\midrule
InternVL2-8B*& 74.71 & 78.29  \\
InternVL2-8B-R & 75.74 & 78.82  \\
\textbf{InternVL2-8B-CoB} & \textbf{77.14} & \textbf{81.43} \\
\bottomrule
\end{tabular}}
\caption{Zero-shot performance of Doc-CoB.}
\label{info_mp}
\end{center}
\end{table}

\begin{figure}
\centering
\includegraphics[width=1\linewidth]{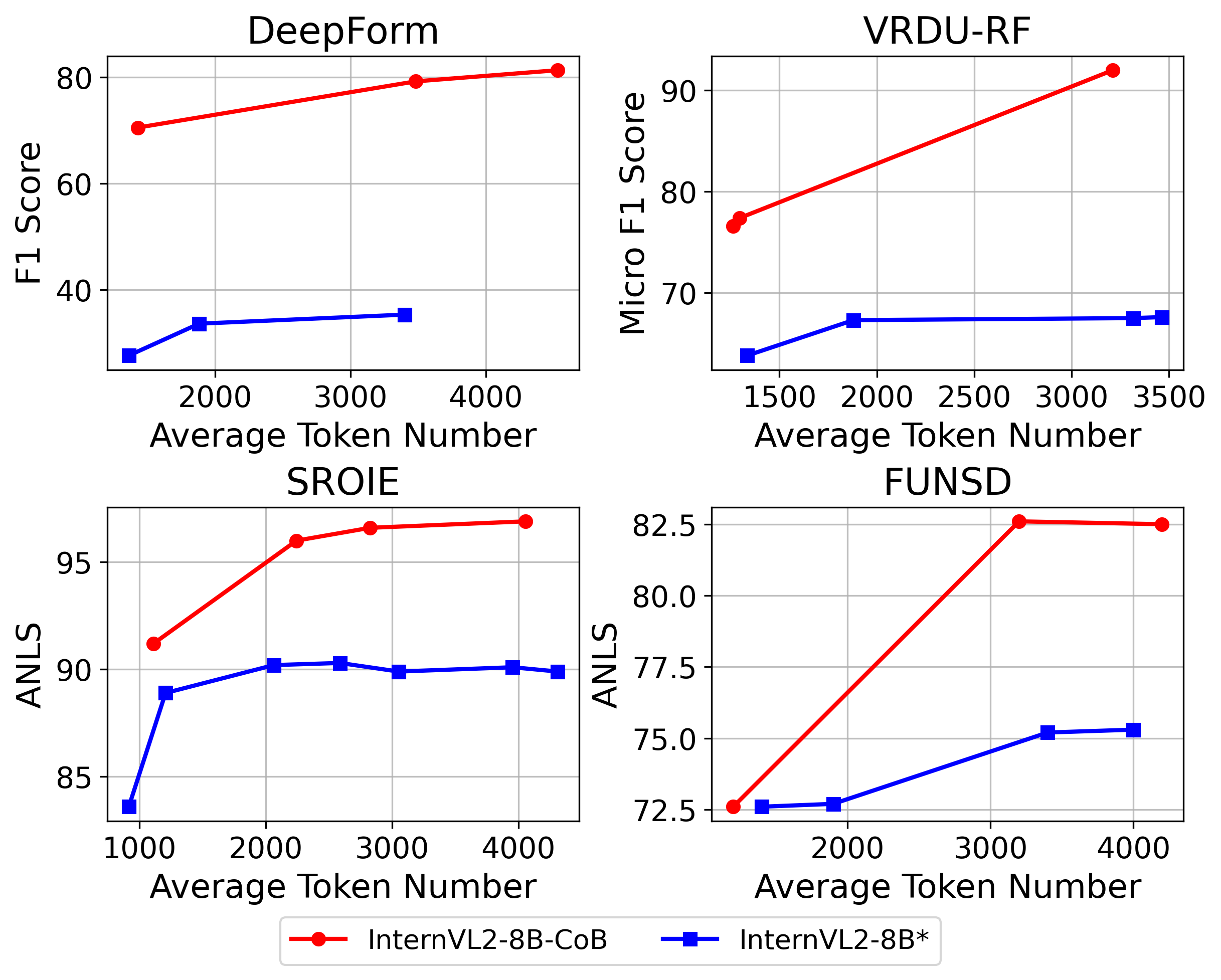}
\caption{Performance of InternVL2-8B* (blue) and InternVL2-8B-CoB (red) on four datasets, plotted against the average token numbers.}
    \label{fig:atask1}
\end{figure}

\begin{table*}[!ht]
\centering
\begin{tabular}{lccccccc}
\toprule
& DeepForm & SROIE & FUNSD & DUDE & VRDU-Ad & VRDU-RF & DocVQA \\ 
\midrule
InternVL2-8B* & 36.33 & 90.39 & 75.17 & 64.12 & 58.66 & 67.58 & 90.27 \\ 
\midrule
\addlinespace[0.5pt]
\multicolumn{8}{c}{\small\textit{\textbf{InternVL2-8B-CoB (layout analyzer variations)}}}\\[-1pt]
 OCR+K-means & 78.56 & 96.82 & 83.97 & 65.03 & 93.91 & 93.42 & 90.77 \\
 Marker & 79.61 & 97.03 & 83.75 & 65.32 & 93.01 & 92.32 & 90.48 \\ 
 MinerU & 80.06 & 97.18 & 82.95 & 65.93 & 93.76 & 92.64 & 91.16 \\
\bottomrule
\end{tabular}
\caption{Performance of InternVL2-8B-CoB with different layout analyzers.}
\vspace{-7pt}
\label{tab:layout_tool}
\end{table*}

\subsection{Robustness to Layout Analyzers}
\label{sec:layout_analyzers}
According to Sec. \ref{sec:reasoning_definition}, Doc-CoB reasoning uses a layout analysis tool to obtain all layout boxes in a given document image. To examine whether Doc-CoB depends on a specific analyzer, we repeat the experiments using two alternative tools. 
The first, we apply the K-means algorithm to spatially cluster OCR-extracted bounding boxes and is referred to as ``OCR + K-means''. The second is Marker\footnote{\url{https://github.com/VikParuchuri/marker}}, an open-source layout analyzer similar to MinerU.
All experiments use the InternVL2-8B-CoB checkpoint, and only the analyzer is changed without retraining the model. The results in Table \ref{tab:layout_tool} show that Doc-CoB consistently improves MLLM performance regardless of the analyzer used. This suggests that Doc-CoB is practical in real-world applications, as any analyzer that provides coarse segmentation is sufficient.

\subsection{Accuracy-Efficiency Trade-off}
\label{tradeoff}
It is widely acknowledged that image tokens constitute the majority of input tokens in MLLMs designed for high-resolution images \citep{chen2024internvl, hu2024mplug}. Doc-CoB reasoning appears to double the total token budget, as two images are processed. However, we hypothesize that Doc-CoB offsets this overhead by guiding the model’s attention to key regions, thereby reducing the effective resolution requirement. 

To validate this, we conduct additional experiments using InternVL2-8B* and InternVL2-8B-CoB, adjusting the hyperparameters that control image resolution (i.e., image tokens). For InternVL2-8B*, the token count includes tokens used for vanilla inference, while for InternVL2-8B-CoB, it includes the total input tokens across both stages. As shown in Figure~\ref{fig:atask1}, Doc-CoB consistently outperforms vanilla inference while using an equal or even fewer tokens. 
Although two-stage inference introduces additional latency, accuracy is often more important in practical document understanding tasks such as compliance review. Therefore, Doc-CoB’s performance gains justify the moderate additional inference cost.

\subsection{Case Study}
\label{sec:casestudy}
\begin{figure}
\centering
\resizebox{\columnwidth}{!}{%
\includegraphics[width=\linewidth]{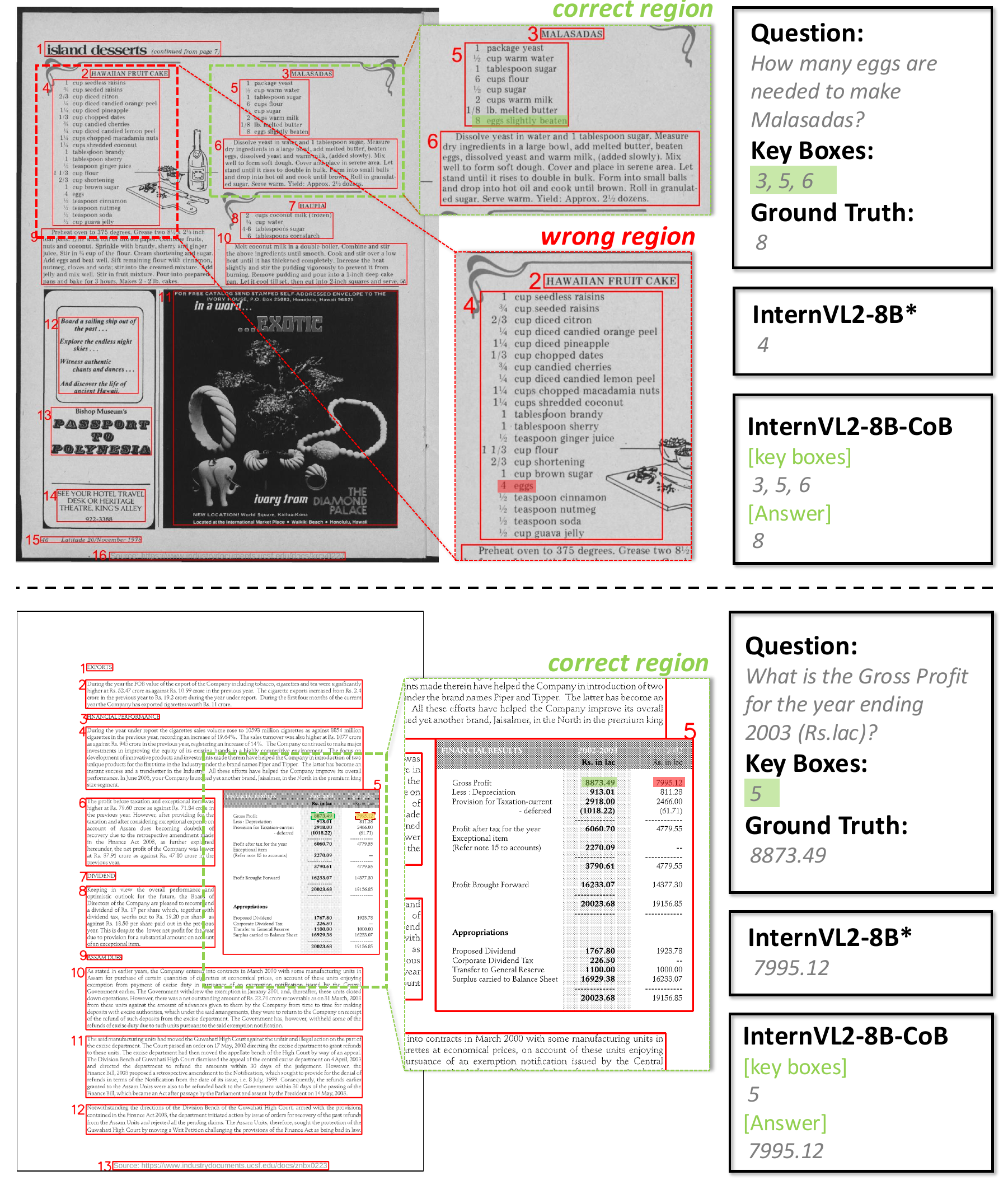}}
\caption{Top: Successful case. Bottom: Failure case.}
\label{fig:casestudy}
\vspace{-7pt}
\end{figure}

Figure \ref{fig:casestudy} presents two cases generated by InternVL2-8B* and InternVL2-8B-CoB (Doc-CoB).
In the top example, the document image contains substantial irrelevant information. InternVL2-8B* outputs \textit{``4''}. In contrast, InternVL2-8B-CoB suppresses the noise by selecting only the key boxes and outputs the correct answer \textit{``8''}.
The bottom example exposes a limitation of Doc-CoB, that currently provides no further granularity to resolve intra-box ambiguity.
More cases and failure analysis are in the Appendix \ref{ap8}.

\section{Conclusion}
\label{sec:conclusion}

In this paper, we propose Doc-CoB, a coarse-to-fine layout-aware document understanding framework that leverages visual reasoning without modifying model architecture. 
By formulating cognitive processes into a two-stage MLLM reasoning paradigm, Doc-CoB significantly enhances the document understanding capabilities of MLLMs.
We also create a tailored dataset with intermediate visual reasoning supervision and introduce two reasoning tasks to support our approach.
Experimental results demonstrate that our method not only outperforms strong baselines but is also transferable across different MLLMs, exhibiting strong generalization capabilities in real-world application. 

\section*{Limitations}
\label{sec:limitation}
While multiple strategies are adopted during data generation to suppress noise introduced by layout analysis, such noise cannot be fully eliminated at inference time and may accumulate across stages.
In addition, the current pipeline performs only a single round of box selection; introducing a self-verification mechanism that iteratively refines the selected key boxes may further improve robustness and performance.

\section*{Acknowledgments}
This work is supported by the National Natural Science Foundation of China (Grant No: 62372408), Ningbo Natural Science Foundation (Grant No: 2025J001).

\bibliography{custom}
\appendix
\label{sec:appendix}
\section{Doc-CoB Stage 1 Prompt}
\label{ap2}
\begin{tcolorbox}[colback=gray!10, colframe=black, arc=5pt, boxrule=1pt]
Which red boxes should be paid attention to when answering the following question: \{\textit{question}\}? 
Output the box ID near the red box.
\end{tcolorbox}

\section{Key Boxes Labeling Prompt and Quality Assurance}
\label{ap1}

\begin{tcolorbox}[colback=gray!10, colframe=black, arc=5pt, boxrule=1pt, breakable]
You are presented with an image containing multiple pre-labeled boxes, each identified by a unique number. You will receive a series of question–answer (QA) pairs. For each question, determine which labeled boxes in the image contain the information needed to arrive at the given answer, adhering to the following rules:
\vspace{\baselineskip}

1. **If the Number of Boxes in the Image Exceeds 3, Output at Least Three Boxes** 

For each question, list all boxes that are genuinely helpful to answer the question. If the number of truly helpful boxes is less than three, please output several boxes that are most likely to cause confusion in answering the question to ensure that at least three boxes are output.
\vspace{\baselineskip}

2. **If the Number of Boxes in the Image is At Most 3, Output Only the Boxes Helpful for Answering the Question** 

For each question, list all boxes that are genuinely helpful to answer the question. Do not output boxes that might cause confusion.
\vspace{\baselineskip}

3. **Output Reason and Content** 

After listing the boxes, for each box, output the reasons (from semantic, layout, etc. perspectives) why it helps or doesn't help answer the question.
\vspace{\baselineskip}

4. **Other Details**

When referring to any box, use the notation \textless box\textgreater num \textless /box\textgreater.
\vspace{\baselineskip}

5. **Output Format** 

For each QA pair, output only the question ID (Q1, Q2, etc.). Output in this strict format:
\vspace{\baselineskip}

Q1: 

HELPFUL BOX: [\textless box\textgreater num\textless /box\textgreater(s)]

CONFUSING BOX: [\textless box\textgreater num\textless /box\textgreater(s)]

Reason for \textless box\textgreater num\textless /box\textgreater: **30-50 word explanation** 
\vspace{\baselineskip}

Q2:

HELPFUL BOX: [\textless box\textgreater num\textless /box\textgreater(s)]

CONFUSING BOX: [\textless box\textgreater num\textless /box\textgreater(s)]

Reason for \textless box\textgreater num\textless /box\textgreater: **30-50 word explanation** 
\vspace{\baselineskip}

Below is an example of the exact format expected:
\vspace{\baselineskip}

Q1:
HELPFUL BOX: [\textless box\textgreater16\textless /box\textgreater]

CONFUSING BOX: [\textless box\textgreater15\textless /box\textgreater, \textless box\textgreater19\textless /box\textgreater]

Reason for \textless box\textgreater16\textless /box\textgreater: **30-50 word explanation** 

Reason for \textless box\textgreater15\textless /box\textgreater: **30-50 word explanation** 

\vspace{\baselineskip}
Q2: 

HELPFUL BOX: [\textless box\textgreater2\textless /box\textgreater, \textless box\textgreater3\textless /box\textgreater, \textless box\textgreater4\textless /box\textgreater]

CONFUSING BOX: []

Reason for \textless box\textgreater2\textless /box\textgreater: **30-50 word explanation** 

Reason for \textless box\textgreater3\textless /box\textgreater: **30-50 word explanation** 

Reason for \textless box\textgreater4\textless /box\textgreater: **30-50 word explanation** 

\vspace{\baselineskip}

**Here are the QA pairs:**

\{QA\_Pairs\}
\end{tcolorbox}

The prompt is utilized for data annotation using GPT-4o, where \textit{QA\_Pairs} represent all question-answer pairs corresponding to the given image. 
The quality assurance process consists of two sequential steps. First, a textual matching rule is applied to validate the format and confirm whether the ground truth appears within the target helpful bounding boxes.
Second, if the annotation of a \textit{QA\_Pair} does not pass the rule-based filtering stage, we use Qwen2-VL-72B-instruct to classify the remaining data, filtering out samples with obvious reasoning or understanding errors. Samples deemed potentially correct are manually reviewed by four researchers for final inclusion.

\section{Training Dataset Details}
\label{ap3}

\textbf{DeepForm} \citep{borchmann2021due} is a document understanding dataset designed for structured information extraction from visually rich documents. It focuses on modeling the hierarchical and relational structure of document elements such as text blocks, tables, and key–value pairs. The dataset supports research on form parsing, layout-aware representation learning, and joint modeling of visual, textual, and structural information.

\textbf{SROIE} \citep{huang2019icdar2019} targets key information extraction from scanned receipt images. It includes annotations for crucial semantic fields such as company name, date, address, and total amount. SROIE is widely used to evaluate end-to-end document understanding systems that combine OCR, layout analysis, and semantic entity recognition.

\textbf{FUNSD} \citep{jaume2019funsd} is designed for semantic understanding of scanned forms. It provides word-level annotations with labels and explicit linking relationships between entities, enabling research on both entity classification and entity linking. FUNSD is commonly used to benchmark layout-aware language models for form understanding tasks.

\textbf{DUDE} \citep{van2023document} is a large-scale dataset aimed at extracting structured information from diverse document types. It emphasizes robustness to layout variation and visual noise, making it suitable for studying generalizable document information extraction methods. The dataset supports tasks such as key–value extraction and document-level semantic understanding.

\textbf{VRDU} \citep{Wang_2023} is a visually rich document understanding dataset that integrates textual content with layout and visual cues. It is designed to evaluate models on complex document layouts, including multi-column text and mixed content types. VRDU facilitates research on multimodal representation learning for documents by combining vision and language features.

\textbf{DocVQA} \citep{mathew2021docvqa} is a benchmark dataset for visual question answering on document images. It requires models to answer natural language questions based on both the textual content and visual layout of documents. DocVQA is widely used to assess document-level reasoning, cross-modal alignment, and the ability of models to locate and interpret relevant information within documents.
Detailed statistics are provided in Table \ref{tab:trainset_questions_images}, Table \ref{tab:enabling_trainset_questions_images}, Table \ref{tab:testset_questions_images_doccob}.

\begin{table*}[t]
\begin{center}
\begin{tabular}{lccccccc}
\toprule
& \multicolumn{1}{l}{DeepForm} & \multicolumn{1}{l}{SROIE} & \multicolumn{1}{l}{FUNSD} & \multicolumn{1}{l}{DUDE} & \multicolumn{1}{l}{VRDU-Ad} & \multicolumn{1}{l}{VRDU-RF} & \multicolumn{1}{l}{DocVQA} \\ 
\midrule
Images & 813 & 626 & 146 & 5239 & 369 & 1451 & 10194 \\
Questions & 3407 & 2499 & 1676 & 10111 & 1972 & 3508 & 39463 \\
\bottomrule
\end{tabular}
\caption{Statistics of the number of images and questions in the \textbf{original} training datasets across seven datasets. Each column represents a different dataset, and the rows provide counts of images and corresponding questions used in the training phase.}
\label{tab:trainset_questions_images}
\end{center}
\end{table*}

\begin{table*}[t]
\setlength\tabcolsep{3pt}
\begin{center}
\resizebox{\textwidth}{!}
{
\begin{tabular}{lccccccccc}
\toprule
& \multicolumn{1}{l}{DeepForm} & \multicolumn{1}{l}{SROIE} & \multicolumn{1}{l}{FUNSD} & \multicolumn{1}{l}{DUDE} & \multicolumn{1}{l}{VRDU-Ad} & \multicolumn{1}{l}{VRDU-RF} & \multicolumn{1}{l}{DocVQA} & \multicolumn{1}{l}{FeTaQA} & \multicolumn{1}{l}{PubLayNet} \\ 
\midrule
Images & 4220 & 3124 & 1822 & 15350 & 2341 & 4959 & 78626 & 39478 & 7597 \\
Questions & 14191 & 10534 & 5782 & 58525 & 8483 & 19540 & 61832 & 72438 & 27167 \\
\bottomrule
\end{tabular}
}
\caption{Statistics of the number of images and questions in the training datasets used for \textbf{reasoning tasks} across nine datasets. Each column represents a different dataset, and the rows provide counts of images and corresponding questions used in the training phase. This includes both Box Recognition Enhancement and Box Reasoning Enhancement.}
\label{tab:enabling_trainset_questions_images}
\end{center}
\end{table*}

\begin{table*}[t]
\begin{center}
\begin{tabular}{lccccccc}
\toprule
& \multicolumn{1}{l}{DeepForm} & \multicolumn{1}{l}{SROIE} & \multicolumn{1}{l}{FUNSD} & \multicolumn{1}{l}{DUDE} & \multicolumn{1}{l}{VRDU-Ad} & \multicolumn{1}{l}{VRDU-RF} & \multicolumn{1}{l}{DocVQA} \\ 
\midrule
Images & 1626 & 1250 & 292 & 10461 & 736 & 2880 & 20293 \\
Questions & 10221 & 7497 & 5028 & 30255 & 5906 & 10396 & 117619 \\
\bottomrule
\end{tabular}
\caption{Statistics of the number of images and questions in the training datasets used for \textbf{Doc-CoB} across seven datasets. Each column represents a different dataset, and the rows provide counts of images and corresponding questions used in the training phase.}
\label{tab:testset_questions_images_doccob}
\end{center}
\end{table*}

\begin{table*}[t]
\begin{center}
\begin{tabular}{lccccccc}
\toprule
& \multicolumn{1}{l}{DeepForm} & \multicolumn{1}{l}{SROIE} & \multicolumn{1}{l}{FUNSD} & \multicolumn{1}{l}{DUDE} & \multicolumn{1}{l}{VRDU-Ad} & \multicolumn{1}{l}{VRDU-RF} & \multicolumn{1}{l}{DocVQA} \\ 
\midrule
Images & 344 & 347 & 47 & 1295 & 543 & 2181 & 1287 \\
Questions & 1423 & 1388 & 467 & 2551 & 3014 & 5318 & 5188 \\
\bottomrule
\end{tabular}
\caption{Statistics of the number of images and questions in the test datasets across seven datasets. Each column represents a different dataset, and the rows provide counts of images and corresponding questions used in the testing phase.}
\label{tab:Statistics}
\end{center}
\end{table*}

\section{Test Dataset Statistics}
Details are provided in Table \ref{tab:Statistics}.
\label{ap4}

\section{GPT-4o and InternVL2-40B Evaluation Prompt}
\label{ap6}
Below is the prompt for InternVL2-40B on \textbf{all} datasets, the prompt for GPT-4o on the \textbf{Document QA} datasets, and the prompt for GPT-4o on the \textbf{Document IE} datasets.
\begin{tcolorbox}[colback=gray!10, colframe=black, arc=5pt, boxrule=1pt]
\{\textit{question}\} Answer the question using a single word or phrase.
\end{tcolorbox}

\begin{tcolorbox}[colback=gray!10, colframe=black, arc=5pt, boxrule=1pt]
You are provided with an image of a document and a question related to it.
Please carefully read the content of the document and answer the question based solely on the information in the image.
The answers to questions are short text spans.
\vspace{\baselineskip}

Question: \{\textit{question}\}
\vspace{\baselineskip}

Directly extract the answer of the question from the document with few words.
\vspace{\baselineskip}

Answer:
\end{tcolorbox}

\begin{tcolorbox}[colback=gray!10, colframe=black, arc=5pt, boxrule=1pt]
You are asked to answer questions asked on a document image.
The answers to questions are short text spans taken verbatim from the document.
This means that the answers comprise a set of contiguous text tokens present in the document.\vspace{\baselineskip}

Question: \{\textit{question}\}
\vspace{\baselineskip}

Directly extract the answer of the question from the document with few words.
\vspace{\baselineskip}

Answer:
\end{tcolorbox}

\section{Failure Analysis and Cases of Doc-CoB}
\label{ap8}
Below is our failure analysis, which resulted in two types of errors.
\textbf{Hallucination Errors (majority):} The MLLM misinterprets the question or struggles to understand the document content, resulting in unrelated outputs.
\textbf{OCR Recognition Errors (minority):} Due to handwritten fonts, image blurriness, or other issues, the model’s perception fails, and it cannot correctly output the textual answer.
Additionally, we observe that InternVL2-8B-CoB corrects 64\% of the errors made by InternVL2-8B, and the performance gains of CoB primarily stem from correcting inter-box errors, rather than improving intra-box accuracy. These case statistics align with the main design goal of Doc-CoB. CoB is intended to enhance the MLLM’s coarse-to-fine reasoning ability at the layout level, rather than at the word-level OCR granularity.

We observed that CoB exhibits minimal performance improvement on DUDE and DocVQA, but achieves significant gains on DeepForm; therefore, we conducted a detailed analysis of the case and found that most of the errors on DUDE and DocVQA are not related to key boxes selection. In other words, the model correctly selected the box ID in the first stage, but due to the noise or complex content within the box, it caused the model to misinterpret the information, resulting in incorrect answers.
Specifically, among all incorrect cases of DocVQA, only 14.98\% involve wrong box selection, compared to 17.38\% (DUDE), 21.74\% (VRDU-RF), and 29.58\% (DeepForm). 
This helps explain why the model shows only a small improvement over the baseline on DocVQA (+0.89), a larger improvement on DUDE (+1.81), and substantial improvements on the other five datasets (averaging +23.69).

\begin{figure*}[t]
    \centering
    \includegraphics[width=0.9\linewidth]{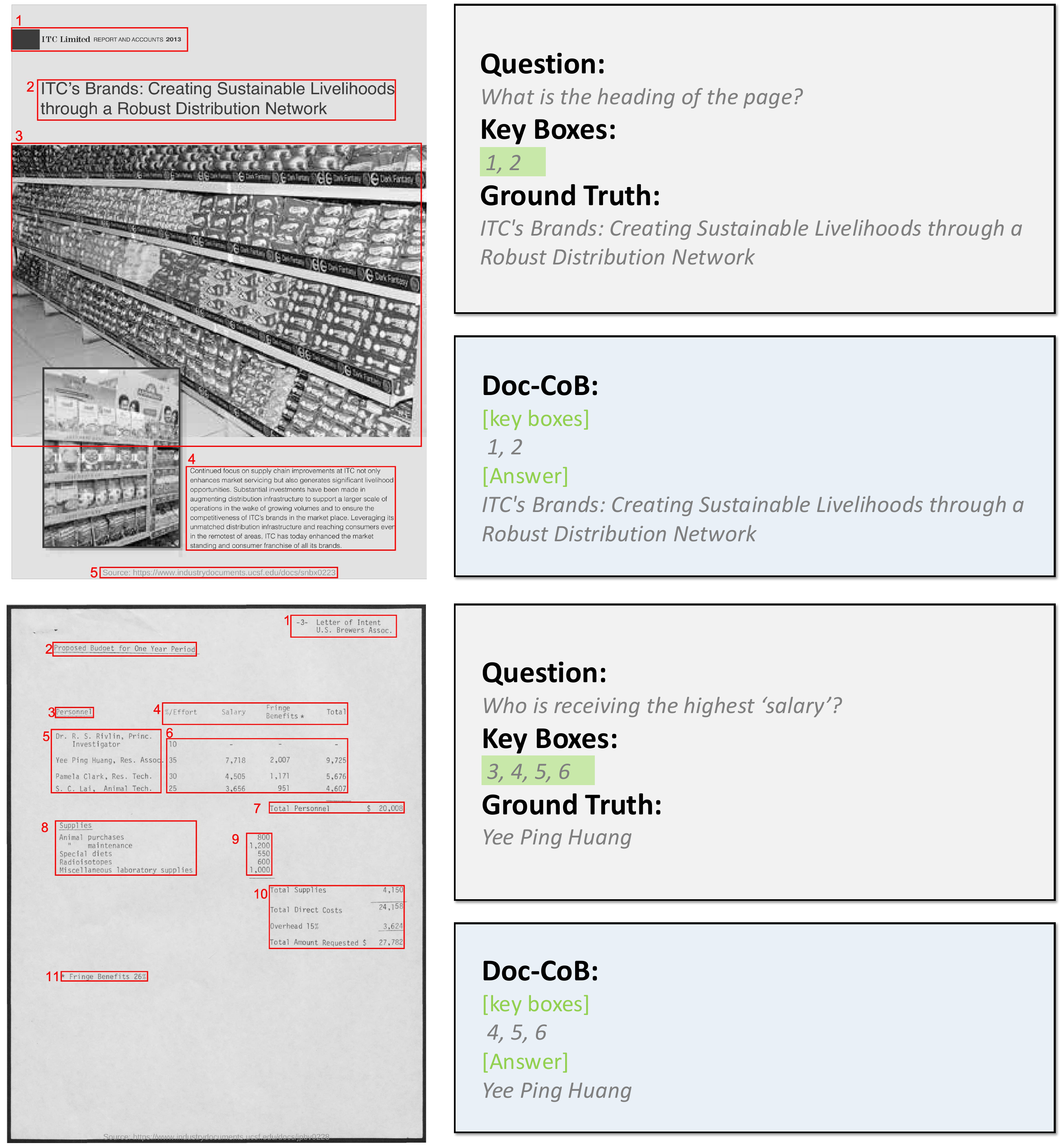}
    \caption{Cases showing intermediate and final outputs of Doc-CoB.}
    \label{fig:enter-label1}
\end{figure*}

\begin{figure*}[h]
    \centering
    \includegraphics[width=0.9\linewidth]{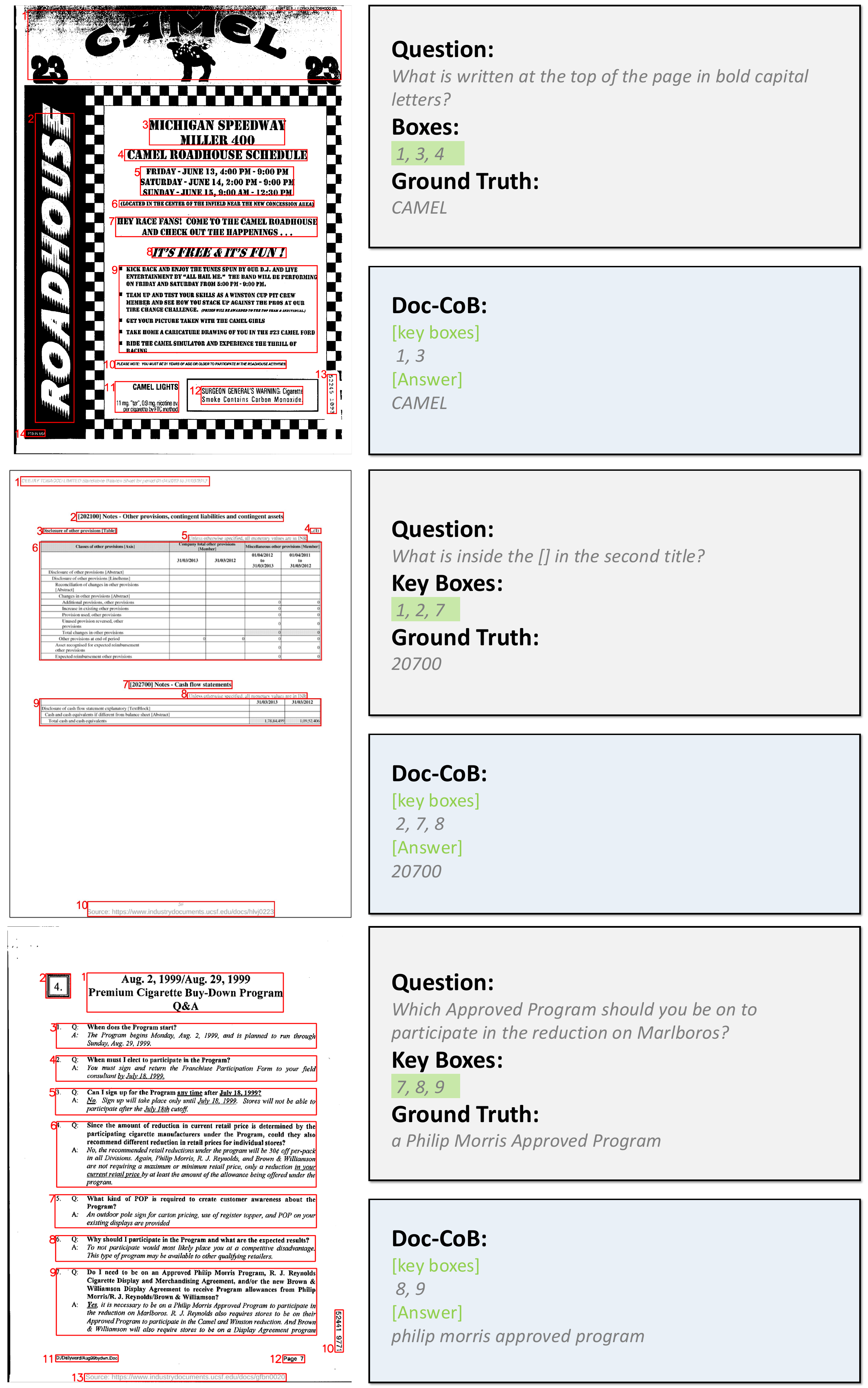}
    \caption{Cases showing intermediate and final outputs of Doc-CoB.}
    \label{fig:enter-label2}
\end{figure*}

\begin{figure*}[h]
    \centering
    \includegraphics[width=0.9\linewidth]{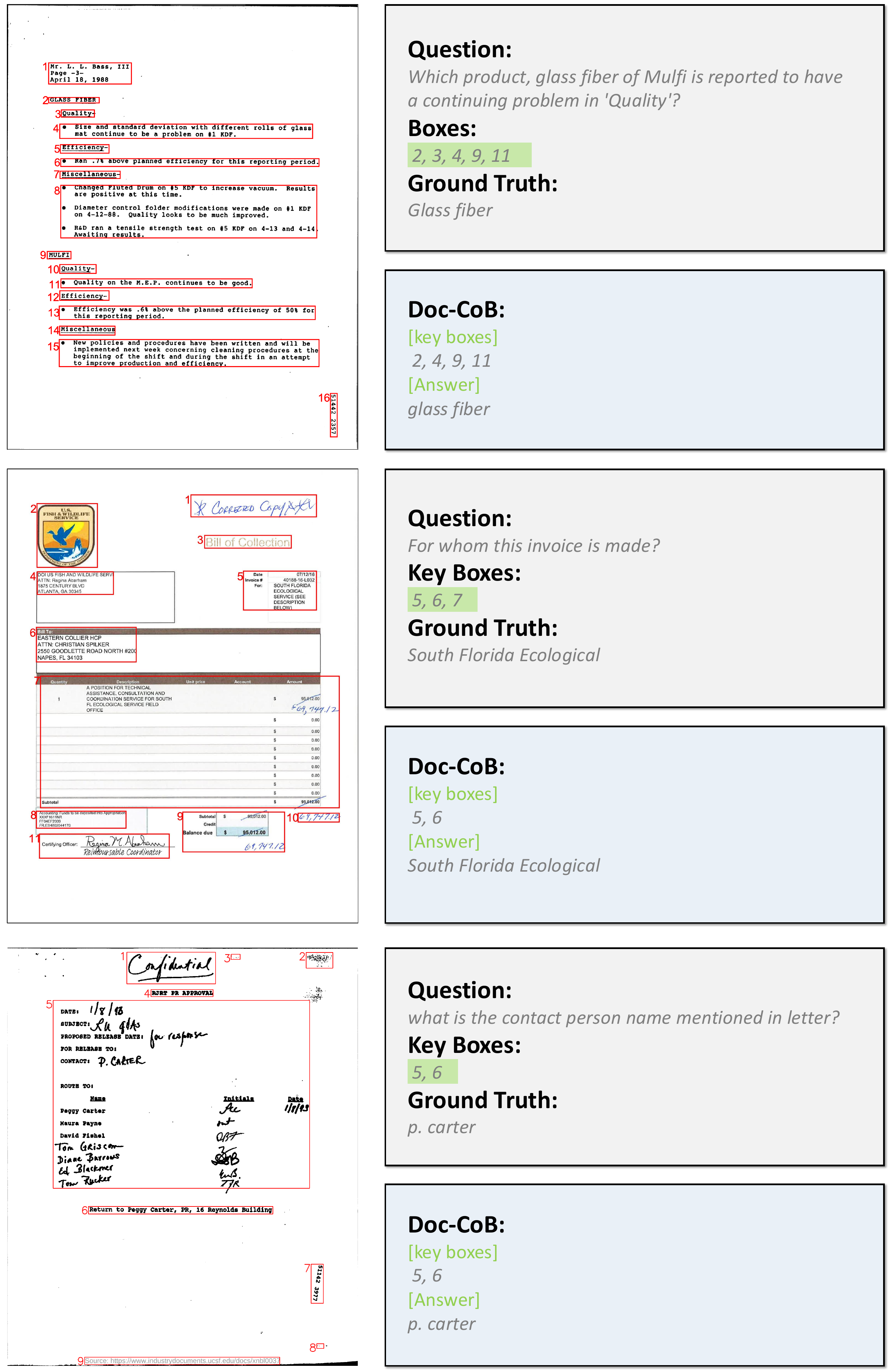}
    \caption{Cases showing intermediate and final outputs of Doc-CoB.}
    \label{fig:enter-label3}
\end{figure*}

\begin{figure*}[h]
    \centering
    \includegraphics[width=0.9\linewidth]{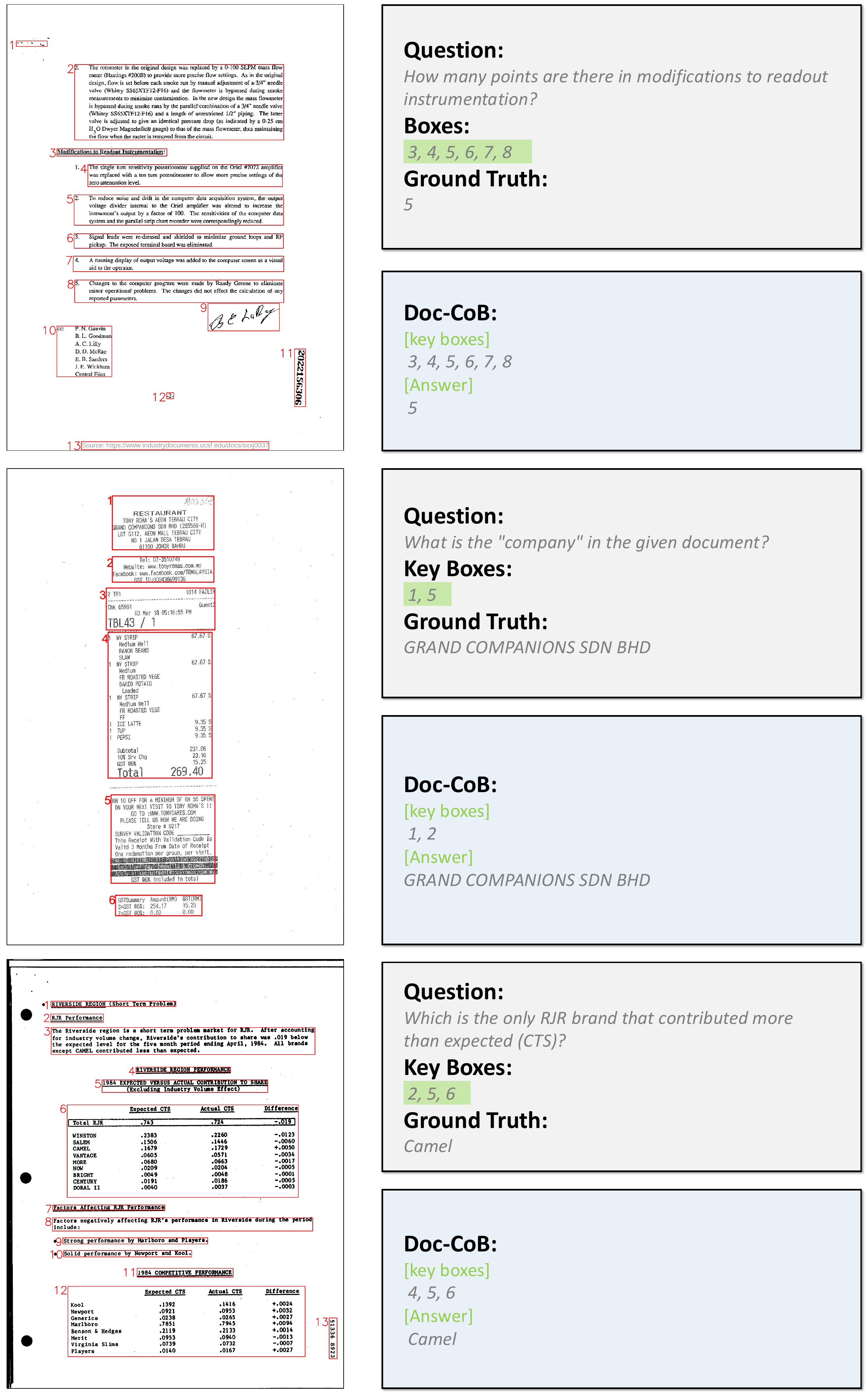}
    \caption{Cases showing intermediate and final outputs of Doc-CoB.}
    \label{fig:enter-label4}
\end{figure*}

\end{document}